\pdfoutput=1

\documentclass[11pt]{article}

\usepackage{acl}

\usepackage{times}
\usepackage{latexsym}
\usepackage{graphicx}
\usepackage{comment}
\usepackage{url}
\usepackage{float}

\usepackage[T1]{fontenc}

\usepackage[utf8]{inputenc}

\usepackage{microtype}
\usepackage{booktabs}
\usepackage{multirow}

\newcommand\todo[1]{\textcolor{red}{\fbox{TODO} #1}}
\newcommand\cl[1]{\textcolor{green}{#1}}

\newcommand\zh{\textsc{zh}\,}
\newcommand\de{\textsc{de}\,}
\newcommand\en{\textsc{en}\,}

\definecolor{forestgreen}{RGB}{34,139,34}
\title{BenchIE: A Framework for Multi-Faceted Fact-Based \\ Open Information Extraction Evaluation}

\author{Kiril Gashteovski$^{1}$, Mingying Yu$^{1, 2}$, Bhushan Kotnis$^1$, Carolin Lawrence$^1$, \\  \textbf{Mathias Niepert}$^{1,4}$, \textbf{Goran Glava\u{s}}$^{2,3}$ \\
  $^1$NEC Laboratories Europe GmbH, Heidelberg, Germany \\
  $^2$University of Mannheim, $^3$LMU Munich, Germany \\
  $^4$University of Stuttgart, Germany \\
  \texttt{firstname.lastname@neclab.eu} \\
  \texttt{goran@informatik.uni-mannheim.de}}

\begin{document}
\maketitle
\begin{abstract}
Intrinsic evaluations of OIE systems are carried out either manually---with human evaluators judging the correctness of extractions---or automatically, on standardized benchmarks. The latter, while much more cost-effective, is less reliable, primarily because of the \textit{incompleteness} of the existing OIE benchmarks: the ground truth extractions do not include all acceptable variants of the same fact, leading to unreliable assessment of the models' performance. Moreover, the existing OIE benchmarks are available for English only. In this work, we introduce BenchIE: a benchmark and evaluation framework for comprehensive evaluation of OIE systems for English, Chinese, and German. In contrast to existing OIE benchmarks, BenchIE is \textit{fact-based}, i.e., it takes into account informational equivalence of extractions:~our gold standard consists of \textit{fact synsets}, clusters in which we exhaustively list all acceptable surface forms of the same fact. Moreover, having in mind common downstream applications for OIE, we make BenchIE \textit{multi-faceted}; i.e., we create benchmark variants that focus on different facets of OIE evaluation, e.g., compactness or minimality of extractions.   
We benchmark several state-of-the-art OIE systems using BenchIE and demonstrate that these systems are significantly less effective than indicated by existing OIE benchmarks. We make BenchIE (data and evaluation code) publicly available.\footnote{\url{https://github.com/gkiril/benchie}}
\end{abstract}

\section{Introduction}



Open Information Extraction (OIE) is the task of extracting relations and their arguments from natural language text in a schema-free manner \cite{banko2007open}. Consider the sentence
\emph{"Sen.~Mitchell, who is from Maine, is a lawyer."}; an OIE system is expected to extract the triples \emph{("Sen.~Mitchell"; "is from"; "Maine")} and \emph{("Sen.~Mitchell"; "is"; "a lawyer")} from the sentence. OIE systems are used in many downstream tasks, including knowledge graph (KG) population \cite{gashteovski2020aligning}, open link prediction 
\cite{broscheit2020}, and question answering \cite{yan2018assertion}. These downstream tasks lend themselves as natural setups for extrinsic OIE evaluation \cite{mausam2016open}.
While valuable in concrete applications, such extrinsic evaluations do not measure the \emph{intrinsic correctness} of the extracted facts: for that purpose, several benchmarks for intrinsic OIE evaluation have been proposed \cite{Stanovsky2016EMNLP,lechelle2019wire57,bhardwaj2019carb}.

Automated benchmark evaluations are more feasible (i.e., faster and cheaper) than manual OIE evaluations \cite{hohenecker2020systematic}. The current benchmarks, however, use scoring functions that are based on approximate (token-level) matching of system extractions against ground truth facts, which seems to be substantially less reliable than human judgments of extraction correctness \cite{zhan2020span}. This primarily stems from the \textit{incompleteness} of existing OIE benchmarks: the gold standard extractions do not include \textit{all} acceptable surface realizations of the \textit{same fact}. 
Consider, for example, a sentence from the recent evaluation framework CaRB \cite{bhardwaj2019carb}: \textit{``Sen.~Mitchell is confident he has sufficient votes to block such a measure with procedural actions''}; with the gold triple extraction (\textit{``Sen.~Mitchell''}; \textit{``is confident he has''}; \textit{``sufficient votes to \dots procedural actions''}). Intuitively, a system extraction with a more concise object---(\textit{``Sen.~Mitchell''}; \textit{``is confident he has''}; \textit{``sufficient votes''})---could also be accepted, as it still captures the same core piece of knowledge, and would arguably be valuable in most downstream tasks. 

To account for this, existing benchmarks credit system extractions for per-slot lexical overlap with gold extractions. Such scoring is overly lenient and overestimates the systems' ability to extract correct \textit{knowledge facts}. Consider, e.g., a system extraction (\textit{``Sen. Mitchell''}; \textit{``is confident he has''}; \textit{``procedural actions''}) for the above-mentioned sentence. From the \textit{factual} perspective, this extraction is clearly incorrect  (\textit{Sen.~Mitchell} has \textit{votes}, not \textit{actions}). However, the popular CaRB benchmark with its token-level metrics would judge the extraction as having (1) perfect precision, since all extracted tokens can be found in corresponding slots of a gold extraction and (2) high recall, as all of the gold subject and predicate tokens as well as two gold object tokens (\textit{``procedural''} and \textit{``actions''}) are found within corresponding slots of the system extraction (Table \ref{tab:carb-benchie-extractions2}). Moreover, by providing a single ground truth extraction per fact, existing OIE benchmarks fail to acknowledge that different downstream applications focus on different facets (i.e., aspects) 
of OIE extractions: e.g., for text summarization, one may prefer
\textit{minimal} extractions \cite{ponza2018facts}, whereas knowledge base population benefits from strict correctness of entities in subject and object slots \cite{lin2020kbpearl}.     

\setlength{\tabcolsep}{10.6pt}
\begin{table*}
    \centering
    \footnotesize
        \begin{tabular}{rrlccc}  
            \toprule
            \multicolumn{6}{l}{\textbf{Input sentence:} \emph{"Sen.~Mitchell is confident he has sufficient votes to block such a measure with procedural actions."}} \\
            \multicolumn{6}{l}{\textbf{CaRB golden extraction:} \emph{("Sen.~Mitchell"; "is confident he has"; "sufficient votes to block ...procedural actions")}} \\
            \cmidrule(r){1-6}
            & \multicolumn{2}{c}{OIE extraction}    & \multicolumn{2}{c}{CaRB (P / R)} & BenchIE \\ \midrule
            $t_1$ & \emph{("Sen.~Mitchell"; "is confident he has";} & \emph{ "sufficient")} & 1.00 & 0.44 & 0 \\ 
            $t_2$ & \emph{("Sen.~Mitchell"; "is confident he has";} & \emph{ "sufficient actions")} & 1.00 & 0.50 & 0 \\
            $t_3$ & \emph{("Sen.~Mitchell"; "is confident he has";} & \emph{ "sufficient procedural actions")} & 1.00 & 0.56 & 0 \\
            \cmidrule(r){1-6}
            $t_4$ & \emph{("Sen.~Mitchell"; "is confident he has";} & \emph{ "sufficient votes")} & 1.00 & 0.50 & 1 \\ 
            \bottomrule
        \end{tabular}
    
    \caption{Difference in scores between CaRB and BenchIE. For the input sentence, CaRB provides only one extraction which covers all the words in the sentence. 
    Then, for each input OIE extraction (from $t_1$ to $t_4$) it calculates token-wise precision and recall scores w.r.t.~the golden annotation. In contrast, 
    BenchIE provides 46 gold extractions for the same sentence and recognizes OIE extractions as valid if they exactly match any of them.} 
    \label{tab:carb-benchie-extractions2}
\end{table*}

In this work, we depart from lenient OIE evaluations based on per-slot token overlaps and propose BenchIE, a novel \textit{fact-centric} and \textit{multi-faceted} OIE evaluation framework and benchmark at the core of which is the following question: 

\vspace{0.5em} 

\noindent \textit{Does the system extraction express the same fact (i.e., the same unit of knowledge) as any of the ground truth extractions (and vice versa) w.r.t. the specific aspect of the OIE extraction that is of interest for one or more downstream applications?}

\paragraph{Contributions.}
BenchIE advances the state of the art in OIE evaluation in the following: \textbf{(1)} it is the first \textbf{fact-centered} approach to OIE evaluation: to reliably answer the above question, we exhaustively list all correct extractions of the same fact. In contrast to existing benchmarks, BenchIE specifies \textit{complete} sets of fact-equivalent extractions (dubbed \textit{fact synsets}), allowing us to avoid error-prone evaluation based on token overlap measures; \textbf{(2)} BenchIE is the first \textbf{multi-faceted} OIE benchmark, allowing to test systems for different aspects of OIE extractions that may be relevant in concrete downstream applications; \textbf{(3)} BenchIE is a \textbf{multilingual} benchmark, covering English, Chinese, and German, and to the best of our knowledge the first with manually annotated (i.e., gold standard) extractions in all languages;\footnote{\newcite{ro2020multi2oie} introduce a multilingual version of the CaRB dataset by machine translating both sentences and extractions. However, automated translation seems to be highly unreliable for OIE -- as shown by \newcite{kotnis2021integrating}, up to 70\% of sentence or extraction translations obtained this way were incorrect.} \textbf{(4)} finally, as a fact-based and multi-faceted benchmark, BenchIE allows us to perform what we believe to be the most comprehensive \textbf{profiling and comparative evaluation} of OIE systems. BenchIE portrays fact extraction abilities of six state-of-the-art OIE models much less favorably and points to their limitations that cannot be detected with existing benchmarks.  


\setlength{\tabcolsep}{4.7pt}
\begin{table*}
    \footnotesize
    \def\arraystretch{0.8}
        \begin{tabular}{rrrr}  
            \toprule
            \multicolumn{4}{l}{\textbf{Input sentence:} \emph{"Sen.~Mitchell is confident he has sufficient votes to block such a measure with procedural actions."}} \\
            \midrule
            $f_1$ & \emph{("Sen.~Mitchell" | "he";} & \emph{"is";} & \emph{"confident [he has sufficient ... actions]")} \\
            \midrule
            $f_2$ & \emph{("Sen.~Mitchell" | "he";} & \emph{"is confident he has";} & \emph{"sufficient votes")} \\
                  & \emph{("Sen.~Mitchell" | "he";} & \emph{"is confident he has";} & \emph{"suff.~votes to block [such] [a] measure")} \\
            \midrule
            $f_3$ & \emph{("Sen.~Mitchell" | "he";} & \emph{"is confident he has sufficient votes to block"} & \emph{"[such] [a] measure")} \\
                  & \emph{("Sen.~Mitchell" | "he";} & \emph{"is confident he has ... to block [such]";} & \emph{"[a] measure")} \\
                  & \emph{("Sen.~Mitchell" | "he";} & \emph{"is confident he has ... to block [such] [a]";} & \emph{"measure")} \\
            \midrule
            $f_4$ & \emph{("Sen.~Mitchell" | "he";} & \emph{"is confident he has ... [such] [a] measure with";} & \emph{"procedural actions")} \\
                  & \emph{("Sen.~Mitchell" | "he";} & \emph{"is confident he has ... [such] [a] measure";} & \emph{"with procedural actions")} \\
            \bottomrule
        \end{tabular}
        \vspace{-0.5em}
         \caption{An example sentence with four BenchIE fact synsets ($f_1$--$f_4$). BenchIE accounts for entity coreference and accepts triples with both \emph{"Sen.~Mitchell"} and \emph{"he"} as subjects: the delimiter ``|'' is just a shorthand notation for different extractions. Similarly, the square brackets ([]) represent a shorthand notation for multiple extractions: triples both with and without the expression(s) in the brackets are considered correct.} 
    \label{tab:benchie-example}
    \vspace{-1em}
\end{table*}

\section{Matching Facts,\,Not Tokens}\label{sec:benchie}

Most OIE systems extract \emph{(subject, predicate, object)} triples, with concepts as subjects and objects and verb 
phrases (VPs) as predicates \cite{banko2007open,stanovsky2018supervised,lauscher2019minscie,gashteovski2017minie,Gashteovski2019OPIECAO}, though systems producing n-ary \cite{akbik2012kraken}, nested \cite{bhutani2016nestie}, and noun-mediated extractions \cite{yahya2014renoun} also exist. Here we follow the most common practice and focus on VP-mediated facts. 
Our novel fact-based benchmark and evaluation paradigm can, however, equally be applied to other types of extractions (e.g., \citet{friedrich2021annie} used this fact-based concept for OIE to create gold annotations for \emph{NE-Centric OIE triples}; i.e., triples where each argument is a named entity and the relations could be either verb phrases or noun phrases).
\subsection{Fact Synsets}
%
We introduce the general concept of a \textit{fact synset}: a set of \textit{all} possible extractions (i.e., different surface forms) for a given fact type (e.g., VP-mediated facts) that are instances of the same fact. E.g., 
given the input sentence from Table \ref{tab:benchie-example}, the 
extractions (\textit{``Sen.~Mitchell''}; \textit{``has sufficient votes to block''}; \textit{``such a measure''}) and (\textit{``Sen.~Mitchell''}; \textit{``has sufficient votes to block''}; \textit{``measure''}) capture the same fact and thus belong to the same fact synset.

Existing benchmarks fail to exhaustively list all acceptable extractions for the same fact. This is precisely why, in order to avoid penalizing systems for correct extractions that are not exactly the same as the gold triples, they resort to lenient token-based performance measures
prone to two types of errors: (1) they punish correct fact extractions that have limited lexical overlap with the gold extraction of the 
same fact, e.g., (\textit{``Sen.~Mitchell''}; \textit{``is confident he has''}; \textit{``sufficient votes'')} 
vs.~(\textit{``Sen.~Mitchell''}; \textit{``is confident he has''}; \textit{``sufficient votes to \dots procedural actions''}) and 
(2) they reward incorrect extractions that have high lexical overlap with a gold extraction, e.g., 
(\textit{``Sen.~Mitchell''}; \textit{``is confident he has''; ``procedural actions''}) vs.~\emph{(``Sen.~Mitchell''; ``is confident he has''; ``sufficient votes to block\dots with procedural~actions'')}. 

To prevent this, BenchIE relies on \textit{exact matching} of system extractions against the gold fact synsets. Further, some OIE systems (over)generate extractions of the same fact; e.g., \emph{(``Sen.~Mitchell''; 
``has sufficient votes to block''; ``such a measure'')} and \emph{("Sen.~Mitchell"; "has sufficient votes to block"; "measure")}. Existing evaluation procedures do not acknowledge the \textit{fact equivalence} of extractions and consequently reward OIE systems for multiply extracting the same fact. Our evaluation based on fact synsets directly remedies these shortcomings of existing OIE benchmarks.     



\subsection{Annotation Process}
\label{subsec:annotation-process}
%
%
\noindent\textbf{English Benchmark.} To make BenchIE comparable to previous benchmarks, we annotate fact synsets on a subset of sentences from CaRB \cite{bhardwaj2019carb}. Because exhaustive annotation of fact synsets is time consuming, we carried it on 300 (out of 1,200) randomly sampled CaRB sentences. To collect truly exhaustive fact synsets, two expert annotators independently labeled the selected 300 sentences in three rounds. \textbf{(1)} Each annotator first (independently) manually denoted every extraction in which a VP-predicate connects two concepts. The annotator then grouped the fact-equivalent triples into fact synsets.\footnote{We provide the annotation guidelines in Appendix~\ref{app:annotation_guidelines_en}.}
%
%
%
%
%
To speed the annotation process up,
we developed a dedicated web-based annotation tool AnnIE that facilitates the extraction of VP-mediated triples (e.g., we color-code verbs to indicate possible predicate heads) and their clustering into fact synsets;\footnote{We show AnnIE's interface in Appendix \ref{app:annotation-tool}. For further details about the tool, see \citet{friedrich2021annie}.} 
%
%
%
%
\textbf{(2)} The annotators then carefully examined all gold extractions from the original CaRB dataset and added those judged to be correct, yet missing from the manually labeled fact synsets from the previous step;
%
\textbf{(3)} Finally, each annotator compared the extractions of all OIE systems in evaluation (see \S\ref{sec:experiments}) against the BenchIE's fact synsets (i.e., the result of the first two steps). Any system extraction not found in BenchIE was carefully examined and---if judged to be correct---added to the appropriate fact synset.\footnote{Very few extractions were actually added in steps (2) and (3); i.e., there were very few correct extractions (from CaRB gold standard and output of OIE systems) that the annotators missed during manual annotation of fact synsets.} Finally, the two annotators merged their independently created annotations by discussing and jointly resolving the disagreements. The overall annotation effort for the English dataset amounted to 80 hours per annotator. English BenchIE contains 136,357 unique gold extractions, grouped into 1,350 fact synsets. For comparison, CaRB \cite{bhardwaj2019carb} lists mere 783 gold triples for the same 300 sentences. Table \ref{tab:benchie-example} shows fact synsets for an example sentence.

\vspace{1.2mm}
\noindent\textbf{Inter-Annotator Agreement (IAA).} To validate BenchIE's annotations, we measure the inter-annotator agreement (IAA) between our two expert annotators. To this end, we quantify the agreement via \textit{recall at the fact level} (see \S\ref{subsec:scoring} for further details): for each annotator, we compute their fact-level recall as the percentage of fact synsets of the other annotator they \textit{cover} with their extractions.\footnote{An extraction \textit{covers} a fact synset if it exactly matches any of the synset's (fact-equivalent) gold triples.} We average the fact-level recalls of the two annotators as the IAA score. We observed a high IAA score of $0.79$. Upon manual inspection, we found that the annotators mostly agree on fact-synset level; most of the the disagreements are on extractions level (particularly, from marking the optional tokens within an extraction; see Appendix \ref{app:subsec-opt-tokens} for details about the optional tokens).



%
     





\vspace{1.2mm}
\noindent\textbf{Chinese and German Benchmarks.}
Two bilingual expert annotators -- native in the target language and fluent in English (\en) -- translated the original 300 English sentences to Chinese (\zh) and German (\de), respectively. Then, to collect exhaustive fact synsets in \zh~and \de, they followed the same three annotation rounds described for \S\ref{subsec:annotation-process}. Due to substantial (primarily syntactic) differences compared to \en, we adjusted the annotation guidelines for these languages (see the Appendix \ref{app:annotation_guidelines_zh} and \ref{app:annotation_guidelines_de} for more details). 
%
The statistics (number of fact synsets and extractions) of the \zh~and \de~benchmarks are given in Table \ref{tab:stats}. Compared to \en~ BenchIE, the \zh~benchmark contains significantly fewer fact synsets (994 compared to 1,350) and more than two orders of magnitude fewer extractions. The drastically smaller number of extractions is primarily due to the lack of determiners and articles in Chinese. Their frequent occurrence in English combined with their neutrality w.r.t.~extractions' correctness results in many mutually different yet fact-equivalent extractions. The numbers for German are, expectedly, much closer to those for English. 


\setlength{\tabcolsep}{4pt}
\begin{table}
    \centering
    \def\arraystretch{0.8}
    \footnotesize
        \begin{tabular}{rrrr}  
            \toprule
                        & \#Extractions  & \#Synsets & \#Extr.~/~Synset\\
                        
            \midrule
            CaRB        & 783  & / & / \\
            \midrule
            BenchIE \en & 136,357 & 1,350 & 101.0 \\ 
            BenchIE \de & 82,260  & 1,086 & 75.7 \\
            BenchIE \zh & 5,318   & 994 & 5.4 \\
            \bottomrule
        \end{tabular}
    \vspace{-0.5em}
    \caption{Multilingual BenchIE: Extraction statistics.}
    \label{tab:stats}
    \vspace{-1em}
\end{table}

\subsection{Evaluation Measure}
\label{subsec:scoring}
We assume that BenchIE is (1) \textit{complete}, i.e., that it contains (a) \textit{all} VP-mediated facts expressed in input sentences and (b) for each fact, its every acceptable extraction as well; and (2) \textit{sound}, i.e., that it does not contain any incorrect extraction that would capture a fact not stated in the sentence. 
Such a complete OIE gold standard enables not only a more reliable evaluation of OIE systems by means of exact matching, but also an evaluation at the more meaningful level of knowledge facts, rather than at the level of individual triples. 

Concretely, we consider a system extraction to be correct if and only if it exactly matches some gold extraction from some fact synset. The number of \textit{true positives (TPs)} is the number of fact synsets (i.e., different facts) \textit{covered} by (at least one of the) system extractions. This way, a system that extracts $N$ different triples of the same fact, will be rewarded only once for the correct extraction of the fact. BenchIE's false negatives (FNs) are then, intuitively, fact synsets not covered by any of the system extractions. Finally, each system extraction that does not exactly match any gold triple (from any synset) counts as a false positive (FP). We then compute \textit{Precision}, \textit{Recall}, and $F_1$ score (as the final score) from TP, FP, and FN in standard fashion. 
\section{Multi-Faceted OIE Benchmark}
\label{sec:facets}

\setlength{\tabcolsep}{2pt}
\begin{table*}
    \centering
    \footnotesize
    \def\arraystretch{0.7}
        \begin{tabular}{lrrr}  
            \toprule
            \multicolumn{4}{l}{\textbf{Input sentence:} \emph{"Sen.~Mitchell is confident he has sufficient votes to block such a measure with procedural actions."}} \\
            \midrule
            
            BenchIE-E & \emph{("Sen.~Mitchell" | "he";} & \emph{"is confident he has ... [such] [a] measure with";} & \emph{"procedural actions")} \\
                  \midrule

            BenchIE-C & \multicolumn{3}{l}{\emph{"(Sen.~Mitchell | he) is confident he has sufficient votes to block [such] [a] measure with procedural actions"}} \\ \midrule

            \multirow{2}{5em}{BenchIE-M} & \emph{("Sen.~Mitchell" | "he";} & \emph{"is confident he has sufficient votes to block measure with";} & \emph{"procedural actions")} \\
                  & \emph{("Sen.~Mitchell" | "he";} & \emph{"is confident he has sufficient votes to block measure";} & \emph{"with procedural actions")} \\ 
            
            \bottomrule
        \end{tabular}
        \vspace{-0.5em}
         \caption{Illustration of BenchIE's \textit{facets} for one fact synset ($f_4$ from Table \ref{tab:benchie-example}): all \textit{acceptable} surface realizations under each facet are shown. ``|'' and square brackets have the same shorthand notation purpose as in Table \ref{tab:benchie-example}.} 
         
    \label{tab:benchie-facets-example}
    \vspace{-1em}
\end{table*}

Different downstream applications care about different aspects of OIE extractions. For IE-based text summarization and simplification \cite{ponza2018facts,vstajner2017leveraging}, e.g., triples should be minimal overall, across all slots (i.e., without unnecessary tokens), but the exact token placement across the slots (e.g., if a preposition is in the predicate or object) does not matter. For entity linking and knowledge base population \cite{lin2020kbpearl}, in contrast, the token placement between slots is critical: a token that is not part of an entity, should not be placed into subject or object. Acknowledging this, we create three additional variants of the English BenchIE, referred to as \textit{facets}, each corresponding to one aspect that is relevant in common OIE applications. This effort addresses recent calls for multi-dimensional analysis of NLP systems \cite{ethayarajh2020utility,narayan2021personalized} and is well-aligned with recent efforts that create multi-faceted benchmarks for other NLP tasks \cite{liu2021explainaboard,vath2021beyond} and datasets \cite{xiao2022datalab}.

\subsection{BenchIE-E}
\label{subsec:benchie-ent}
The default, general-purpose BenchIE facet from the previous section was designed to be somewhat tolerant to token distribution accross slots (see Appendix \ref{app:fact-synset} for details): some tokens may be placed in either the predicate or object (e.g., the preposition \textit{with} in the synset $f_4$ in Table \ref{tab:benchie-example}). This enables a more flexible comparison of OIE systems that are designed for different purposes (i.e., systems that produce slightly different token placements are not punished) and is in line with prior work on intrinsic OIE evaluation, both automatic \cite{Stanovsky2016EMNLP,bhardwaj2019carb} and manual \cite{fader2011identifying,del2013clausie,gashteovski2017minie}. 
Such extraction flexibility, however, may not be desirable in tasks like automated KG construction \cite{wolfe2017pocket,jiang2019role} or entity linking \cite{lin2020kbpearl,lin2021tenet}. \newcite{angeli2015leveraging} show empirically that extractions with wholesome entities and without additional tokens yield benefits in KG construction. 

Since OIE is predominantly used for KG-related tasks \cite{weikum2020machine}, it is paramount to have an evaluation facet that imposes strict(er) token boundaries on entity slots -- subjects and objects. We thus create the \textit{entity facet} of the benchmark (BenchIE-E) with this additional constraint of wholesomeness of subject and object concepts. BenchIE-E was constructed by one of our annotators (see \S\ref{subsec:annotation-process}) by removing from \en BenchIE's fact synsets the extractions in which subject and/or object was not a wholesome concept (see Table \ref{tab:benchie-facets-example}). 



\subsection{BenchIE-C}
%
The default BenchIE facet (\S\ref{sec:benchie}) compares OIE extractions against gold triples from fact synsets at the slot level: to be judged correct, an extraction must exactly match some gold triple in all slots. This criterion, however, is overly strict if extractions are to be used in applications like summarization or simplification \cite{ponza2018facts,vstajner2017leveraging}, which commonly concatenate the content of the slots. In this case, it does not matter if a sequence of tokens occurs at the end of the subject or beginning of the predicate (analogously for predicate and object). To reflect this, we introduce the \textit{concatenation facet}, BenchIE-C: for each gold BenchIE triple, we create the gold BenchIE-C utterance by simply concatenating the content of the triple's slots (see Table \ref{tab:benchie-facets-example}). 
\subsection{BenchIE-M}
Our third additional evaluation facet addresses the aspect of \textit{minimality} of OIE extractions \cite{gashteovski2017minie}. More compact extractions can benefit both text generation \cite{ponza2018facts,vstajner2017leveraging} and KG-related tasks \cite{lin2020kbpearl,lin2021tenet}. If two triples $t_1$ and $t_2$ capture the same fact (i.e., are in the same fact synset), $t_1$ is considered \textit{more compact} than $t_2$ if tokens of each $t_1$ slot make a (non-strict) subsequence of tokens in the corresponding $t_2$ slot \cite{gashteovski2020compact}.\footnote{At least one $t_1$ slot has to be a strict subsequence of the respective $t_2$ slot; $t_1$ and $t_2$ would be the same otherwise.} 
%
%
To allow for evaluation of minimality, BenchIE-M triples contain only the non-optional tokens (denoted in square brackets in Table \ref{tab:benchie-example}) from the corresponding BenchIE triple. Consequently, BenchIE-M fact synsets on average contain many fewer extractions than the original BenchIE synsets.\footnote{This does not imply that each fact synset in BenchIE-M contains only one (i.e., minimal) triple (see Table \ref{tab:benchie-facets-example}).}  



\section{Fact-Level Evaluation}
\label{sec:experiments}

\setlength{\tabcolsep}{7pt}
\begin{table*}[!htbp]
    \centering
    \def\arraystretch{0.8}
    \footnotesize
        \begin{tabular}{rrrrrrrrrrr}  
            \toprule
            & & \multicolumn{7}{c}{\textsc{En}} & \textsc{Zh} & \textsc{De} \\
            \cmidrule(l){3-9}\cmidrule(l){10-10} \cmidrule{11-11}
                      &  & Naive OIE & ClausIE & MinIE & Stanford      & ROIE    & OpenIE6   & M$^2$OIE & M$^2$OIE & M$^2$OIE \\
            \midrule
            \multirow{2}{*}{P} 
                    & CaRB & 0.24 & 0.58    & 0.45   & 0.17        & 0.44       & 0.48     & \textbf{0.60}   & /         & /      \\
                & BenchIE  & 0.03 & \textbf{0.50}    & 0.43   & 0.11        & 0.20       & 0.31    & 0.39  & 0.18       & 0.09 \\
                & $\Delta$ & +0.21 &   +0.08    & +0.02  & +0.06  &  \textbf{+0.24} &  +0.17 & +0.21   &  / & / \\
            \midrule
            \multirow{2}{*}{R}      
                & CaRB    & \textbf{0.70} & 0.53    & 0.44   & 0.29         & 0.60       & 0.67        & 0.61    & / &     /   \\
                & BenchIE  & 0.02 & 0.26    & \textbf{0.28}   & 0.16         & 0.09       & 0.21        &  0.16   & 0.10      &  0.03\\
                & $\Delta$ & \textbf{+0.68} & +0.27 & +0.16 & +0.13 & +0.51  & +0.46   & +0.45 & / & / \\
            \midrule
            \multirow{2}{*}{$F_1$}          
                & CaRB   & 0.36 & 0.56    & 0.44   & 0.22         & 0.51       & 0.56   & \textbf{0.61} & /   &  /    \\
            & BenchIE    & 0.03 & \textbf{0.34}    & \textbf{0.34}   & 0.13        & 0.13       & 0.25  & 0.23 & 0.13     &  0.04 \\
            & $\Delta$& +0.33 & +0.22 & +0.10  & +0.09 & \textbf{+0.38}  & +0.31  & \textbf{+0.38} & / & / \\
            \bottomrule
        \end{tabular}
        \vspace{-0.5em}
    \caption{Comparison of performance of OIE systems on BenchIE and CaRB benchmarks for precision (P), recall (R) and $F_1$ score ($F_1$). The row $\Delta$ indicates the difference between CaRB score and BenchIE score (i.e., $\Delta=CaRB-BenchIE$). \textbf{Bold numbers} indicate highest score per row (i.e., highest score for P / R / F$_1$ per benchmark) or highest score difference per row (i.e., highest $\Delta$ for P / R / F$_1$).} 
    \label{tab:experiments}
    \vspace{-1em}
\end{table*}

We first compare BenchIE's fact-level evaluation (i.e., default facet, \S\ref{sec:benchie}) against CaRB's token-level scoring \cite{bhardwaj2019carb}.\footnote{CaRB is an improved version of the widely-adopted OIE2016 benchmark \cite{Stanovsky2016EMNLP}; our findings for CaRB are thus likely to hold for OIE2016 as well.} Our quantitative results confirm our intuitions and observations (see Table \ref{tab:carb-benchie-extractions2}): CaRB systematically and substantially overestimates OIE systems' performance. BenchIE, we argue, portrays the fact extraction abilities of OIE systems more realistically.

\subsection{Experimental Setup}

\paragraph{OIE Systems.} 
\label{subsec:oie-systems}%
We tested six widely used OIE systems that extract VP-mediated facts for \en, namely: ClausIE \cite{del2013clausie}, Stanford OIE \cite{angeli2015leveraging}, 
MinIE \cite{gashteovski2017minie}, ROIE \cite{stanovsky2018supervised}, OpenIE 6 \cite{kolluru2020openie6} and M$^2$OIE \cite{ro2020multi2oie}. We additionally implemented the following naive baseline (Naive OIE): each verb (detected using spaCy's POS-tagger \cite{honnibal2017spacy})
becomes the predicate, its entire preceding sentence context becomes the subject and succeeding context the object. 
%
For \zh and \de, we evaluated a supervised M$^2$OIE \cite{ro2020multi2oie} model based on the multilingual BERT \cite{devlin2019bert}, trained on a large \en dataset \cite{zhan2020span} and transferred (zero-shot) to target languages by means of its multilingual encoder. 

\vspace{1.2mm}
\noindent \textbf{Implicit and N-ary Extractions.} 
Some OIE systems produce implicit extractions containing tokens that do not occur in the sentence.\footnote{E.g., the triple \emph{("Biden"; "be"; "President")} extracted from the phrase \emph{"President Biden ..."}} As BenchIE does not contain implicit annotations, we remove such extractions from the OIE systems' output, to avoid penalizing OIE systems for extracting fact types not covered by the benchmark. 
To make CaRB directly comparable, we automatically remove all its implicit extractions too. ROIE and M$^2$OIE produce N-ary extractions (i.e., more than three slots), whereas BenchIE contains only triples. We follow standard practice \cite{del2013clausie} and convert those extractions into triples by concatenating the third and subsequent slots into a single object. 






\subsection{Results and Discussion}
Table \ref{tab:experiments} summarizes results of OIE systems on BenchIE and CaRB. Across the board, BenchIE's fact-level precision and recall are significantly lower than CaRB's respective precision and recall computed on token level. On average, CaRB scores the OIE systems higher than BenchIE by 14 percentage points for precision, 38 percentage points for recall and 26 percentage points for the $F_1$ score. 

\vspace{1.2mm}
\noindent\textbf{Precision.} 
System's precision on BenchIE is lower (albeit not so drastically lower as recall) than on CaRB because BenchIE, as a complete benchmark, punishes \textit{incorrect facts}, i.e., extractions that cannot be found in BenchIE's fact synsets. CaRB, on the other hand, rewards any token overlap that the incorrectly extracted fact has against its gold triple(s) -- in many cases such overlap is substantial and CaRB consequently rewards the incorrect fact with high precision. Consider, for example, the sentence from Table \ref{tab:carb-benchie-extractions2} and an incorrect fact extraction (\textit{``Sen.~Mitchell''}; \textit{``is confident he has''}; \textit{``sufficient actions''}); on BenchIE, this extraction is a false positive because it does not exist in any of the four fact synsets it lists for the sentence. CaRB, in contrast, rewards the extraction with perfect precision because all its tokens are accounted for in the corresponding slots of its gold triple (\textit{``Sen.~Mitchell''}; \textit{``is confident he has"}; \emph{"sufficient votes to \dots actions''}). 

In an attempt to quantify how much CaRB overestimates fact-level precision with its token overlap metric, we evaluated our Naive OIE baseline on both CaRB and BenchIE. While BenchIE reflects the poor quality of naive extractions with the near-zero performance, CaRB estimates its precision to be non-negligible (0.24) and even higher than that of the Stanford's OIE system (0.17). 
In contrast, BenchIE assigns much lower score to this baseline: precision of 0.03---8 times less than CaRB's score.

\vspace{1.2mm}
\noindent\textbf{Recall.} While CaRB somewhat overestimates fact-level precision of OIE systems, its overestimation of their recall is much more drastic: all tokens of its gold extractions that can be found in respective slots of a factually incorrect extraction of an OIE system contribute to the system's recall. The overestimation of CaRB's recall scores is best illustrated by the fact that our naive baseline (Naive OIE) obtains a score of 0.7, better than any of the six OIE systems under evaluation. In terms of recall, CaRB obviously rewards long extractions -- the longer the system extraction is, the more likely it is to cover more tokens from gold standard extractions. Neural extractors OpenIE6, ROIE, and M$^2$OIE on average produce much longer extractions than rule-based systems like MinIE or Stanford (e.g., on average, a ROIE extraction has 16 tokens, whereas Stanford extraction has 7.7 tokens): accordingly, CaRB rewards the neural systems with much higher recall scores. BenchIE, on the other hand, credits only the OIE extractions that cover its fact synsets (and only once per fact synset). Our Naive OIE is, intuitively, highly unlikely to match gold extractions from fact synsets and BenchIE reflects this with a fact-level recall of only 2\%. Similarly, BenchIE's recall scores reveal that the long extractions of neural OIE systems very rarely correspond to any acceptable variant of an expressed fact (e.g., ROIE's fact-level recall is only 9\%).    





\vspace{1.4mm}
\noindent\textbf{Multilingual OIE.}
We evaluated M$^2$OIE (as the only multilingual model in our evaluation) on the Chinese and German versions of BenchIE. Quite expectedly, the performance for Chinese and German in target languages is below the source English performance. However, the drop due to the zero-shot language transfer is, at first glance -- surprisingly, much larger for German than for Chinese: this goes against findings from other tasks, where transfer performance correlates with linguistic proximity between the source and target language \cite{lauscher2020zero}.  
M$^2$OIE's Chinese performance is encouraging, as it surpasses the English performance of some of the other OIE models 
(e.g., its recall score is better than ROIE, and its precision score is better than Stanford's).
We believe this is because (a) OIE is a highly syntactic task; and (b) Chinese language is syntactically simple and has the same word order as English (SVO). German language, on the other hand, despite overall linguistic proximity to English, has a different word order (SOV; from generative perspective), with the main verb often appearing at the very end of the sentence -- this, we believe, is the main cause of poor OIE transfer between English and German.   
We believe BenchIE is a good starting point for multilingual OIE evaluation: 
we subsequently created additional data for Arabic, Galician, and Japanese:~see~\citet{kotnis2021integrating}~and~\citet{friedrich2021annie} for details and further analyses.

\section{Profiling OIE Systems with BenchIE}
\label{sec:profiling}

Token-based evaluation of existing OIE benchmarks (with real per-extraction scores in the range $[0,1]$) makes pinpointing of extraction error source difficult. This limits their usability in automatic error analysis and system profiling. The fact that previous work performed OIE error analyses manually \cite{fader2011identifying,schneider2017relvis} confirms this. 
BenchIE, in contrast, lists all acceptable extractions and thus naturally lends itself to reliable automatic error analysis and profiling. 



\subsection{Slot Errors}
\label{subsec:slot-errors}
%
%


\begin{figure}
    \centering
    \includegraphics[scale=0.45]{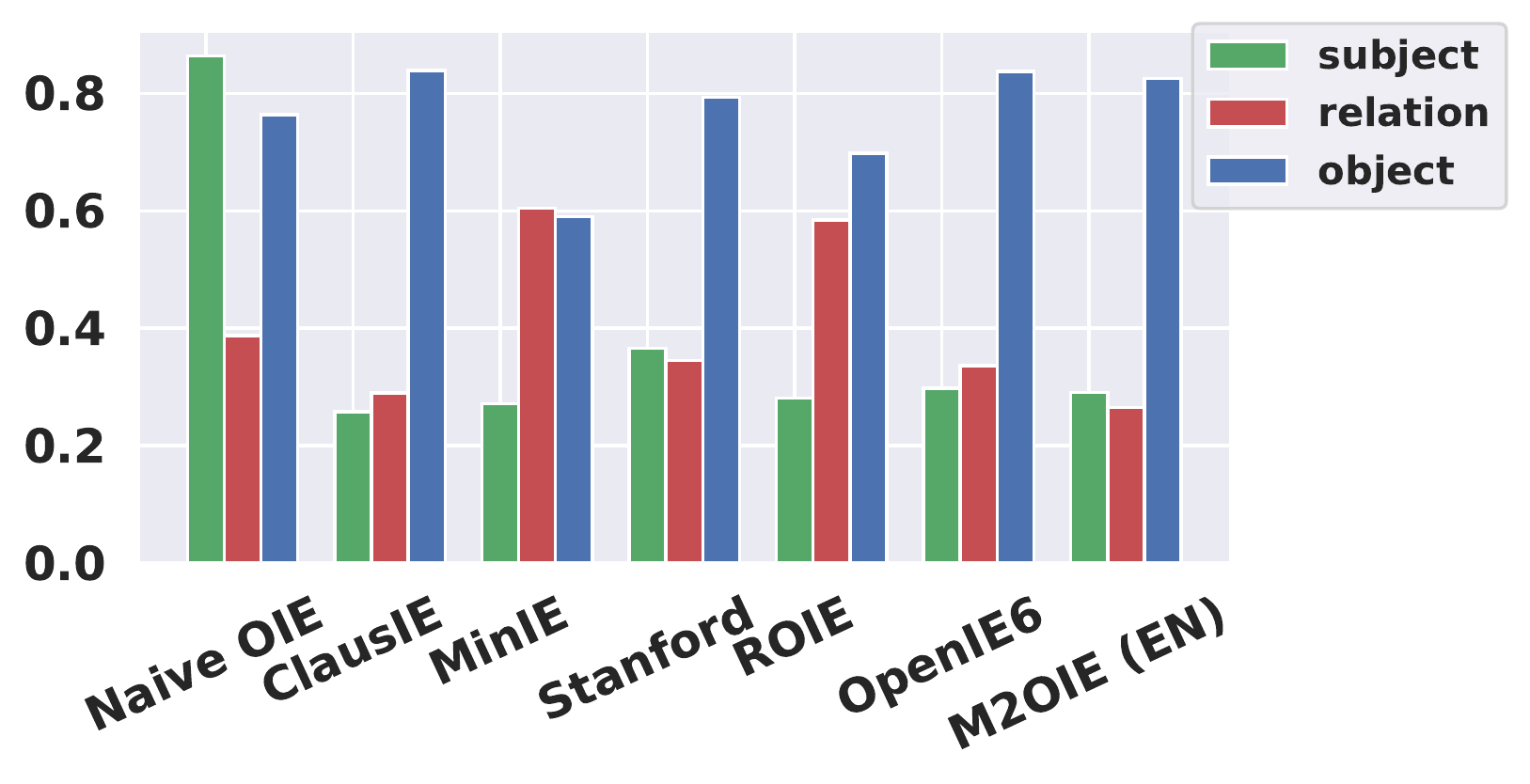}
    \vspace{-0.5em}
    \caption{Relative proportion of errors per slot for OIE systems. Note that (1) fractions do not add up to 1 as extraction can be erroneous in more than one slot; and (2) the figure does not indicate systems' absolute error rates (for performance comparison, see Table \ref{tab:experiments}).}
    \label{fig:slot-errors}
    \vspace{-0.5em}
\end{figure}

\begin{figure}
    \centering
    \includegraphics[scale=0.59]{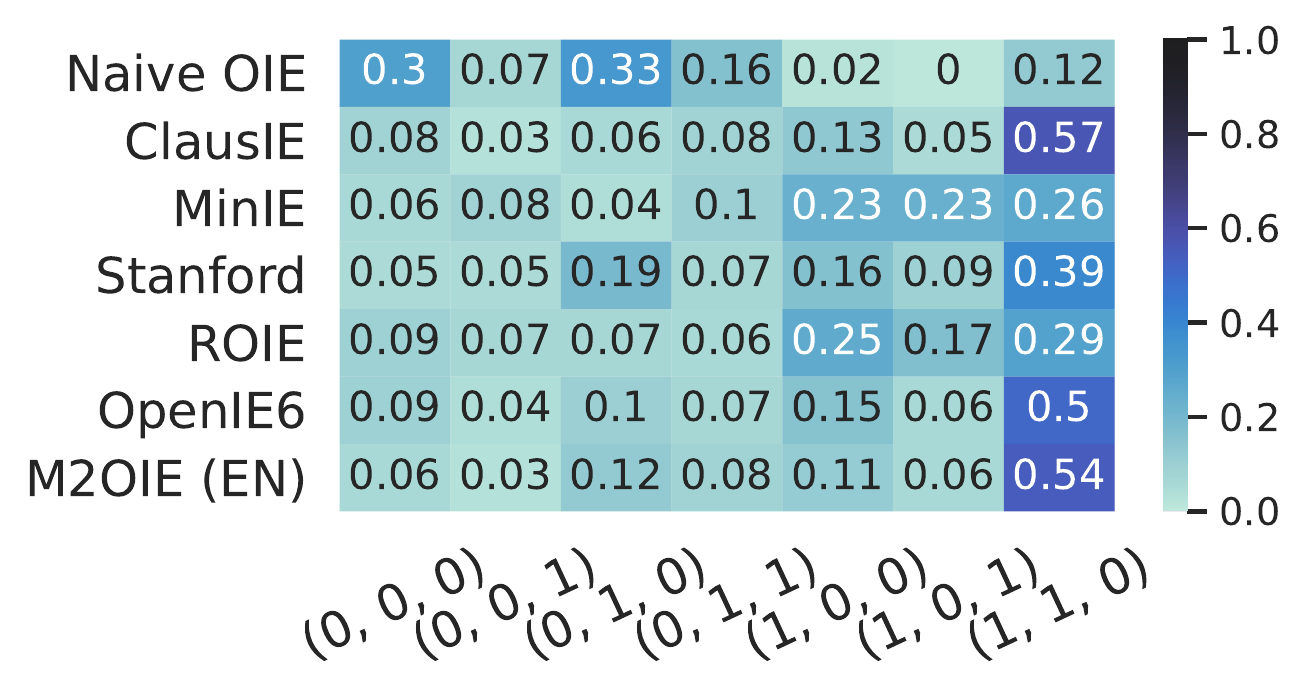}
    \vspace{-2em}
    \caption{Distribution of incorrect extractions of OIE systems across different slot-error combinations.}
    \label{fig:slot-error-dist}
    \vspace{-1em}
\end{figure}


%

We carry out the analysis of errors per slots on the default BenchIE facet (\S\ref{sec:benchie}), because it is application-agnostic, unlike the additional facets from \S\ref{sec:facets}.
We observed that most of the errors in all OIE systems stem from extracting the objects (see Figure \ref{fig:slot-errors}).  
For an SVO language like English, correctly extracting subjects and predicates seems substantially easier than correctly extracting objects.   
MinIE (rule-based) and ROIE (neural) have higher shares of predicate mis-extractions. MinIE post-processes ClausIE's triples by moving words from objects to predicates. Since ClausIE most frequently makes object errors, this effectively redistributes those errors between predicates and objects of MinIE's extractions. 

Figure \ref{fig:slot-errors}, however, does not tell the whole story, as many extractions are erroneous in multiple slots. For more detailed insights, we assign each incorrect extraction to one of seven error buckets: each error bucket indicates one combination of extraction errors across the three slots. For example, the bucket $(1, 1, 0)$ contains extractions that match their \textit{closest} gold triple in the subject and predicate, but not object. The closest gold triple is the one that matches the extraction in most slots.\footnote{An incorrect extraction may have several ``closest'' gold triples that correspond to different error buckets. 
In this case, we increase the count for all competing buckets.} 
The error-bucket analysis, summarized in Figure \ref{fig:slot-error-dist}, reveals that, across all systems, most extractions with object errors actually have correct subjects and predicates (bucket $(1, 1, 0)$). MinIE deviates from this pattern and produces also many extractions with both incorrect object and predicate (bucket $(1, 0, 0)$) or only bad predicate (bucket $(1, 0, 1)$). 
Expectedly, most extractions of our naive baseline most often get only the predicate right (bucket $(0, 1, 0)$) or all three slots wrong (bucket $(0, 0, 0)$). This further emphasizes how misleading current token-based benchmarks can be -- CaRB rewards this baseline with very high recall (see \S\ref{sec:experiments}).       

\begin{figure*}
    \centering
    \includegraphics[scale=0.52]{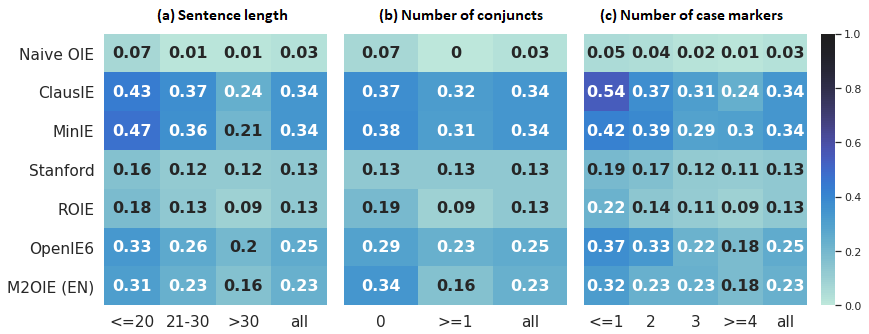}
    \caption{Bucketized experiments: F1 score according to different bucketizations of the input sentences: sentence length (a); number of conjunctions (b); number of case markers (c).}
    \label{fig:buckets-1}
    \vspace{-0.5em}
\end{figure*}

\subsection{Bucketized Error Analysis}

To understand where OIE systems fail systematically, we split the input sentences into buckets and measured the performance of OIE systems per bucket. Based on preliminary qualitative error analysis, we chose bucketization according to some linguistic properties of the sentences that produced erroneous triples. In particular, we examine the performance of OIE systems for sentence length, presence of conjunctions and case markers, since these appeared to be the most common reasons for failure. Note that BenchIE instances can be ``bucketized'' according to an arbitrary dimension interest, lending itself to diverse future fine-grained evaluations and analyses of OIE systems' behaviour. In general, we found that OIE systems exhibit weakest performance on long sentences (with more than 30 tokens) as well as those that contain conjunctions or have more than two case markers (Figure~\ref{fig:buckets-1}). For a more detailed discussion, see Appendix \ref{app:bucket-experiments}.


\subsection{Multi-Faceted Evaluation}

\begin{figure*}
    \centering
    \includegraphics[scale=0.49]{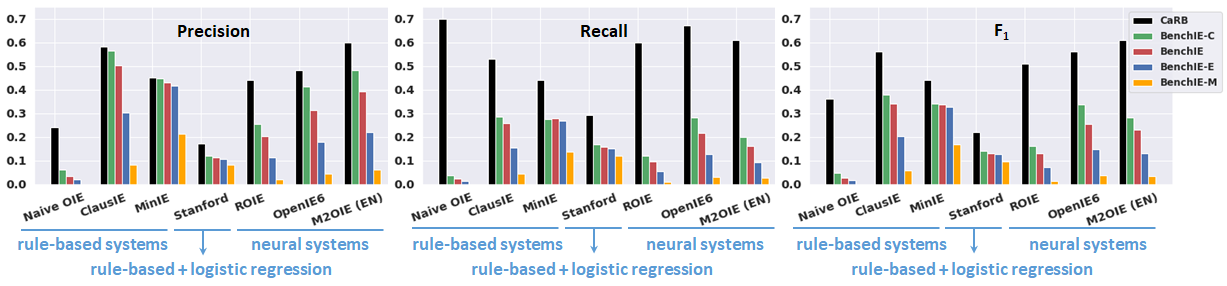}
    \caption{Multi-faceted evaluation of OIE systems.}
    \label{fig:multifaceted-eval-f1}
    \vspace{-1em}
\end{figure*}

Finally, we profile the OIE systems on our three special benchmark facets (\S\ref{sec:facets}): BenchIE-E, -C and -M. Figure \ref{fig:multifaceted-eval-f1} summarizes the performance of OIE systems on these three facets. 




\vspace{1.2mm}

\vspace{1.2mm}
\noindent \textbf{BenchIE-C.} Ignoring slot boundaries, this facet is more lenient to OIE systems than the default facet -- BenchIE-C yields higher scores than the regular BenchIE facet for \textit{all} systems. The gap between the system's performance on BenchIE-C and BenchIE effectively quantifies how often the system misplaces tokens between adjacent slots.  
This gap is very small for Stanford OIE and MinIE -- this means that, for extractions with correct overall token span, they also distribute the tokens between the slots correctly. For downstream tasks like text summarization, BenchIE-C results point to ClausIE as the best choice. 
Interestingly, we observed that CaRB's Precision for some systems (ClausIE and MinIE) effectively matches their Precision on BenchIE-C (see Figure \ref{fig:multifaceted-eval-f1}), which is another indication that CaRB scores, in effect, neglect precise token distributions across slots.

\vspace{1.2mm}
\noindent \textbf{BenchIE-E.} This facet is stricter than the default BenchIE facet -- it allows fewer token placement variants in subject and object. For all OIE systems the $F_1$ BenchIE-E score is thus lower than the corresponding BenchIE score. 
MinIE and Stanford OIE obtain very similar performance on BenchIE-C, BenchIE (default), and BenchIE-E: this means that their extraction (when correct in overall token span) most often have clean concepts in subject and object.
All neural systems and ClausIE exhibit huge performance drops on BenchIE-E -- this means that their subject and object concept extractions are not clean, which makes these systems less suitable for tasks like KG population and entity linking. Out of the systems we evaluate, MinIE is the best fit for such downstream tasks.  

\vspace{1.2mm}
\noindent \textbf{BenchIE-M.} This facet yields the lowest performance for all systems, as it punishes extractions with any unnecessary tokens. 
Expectedly, MinIE -- a system tailored to produce minimal extractions -- yields the best performance on this facet. But even MinIE ``loses'' half of its performance when minimality is enforced (BenchIE vs.~BenchIE-M). This calls for more work on minimizing OIE extractions.  
Stanford OIE outperforms all systems except MinIE, which renders it a good pick when extraction minimality is beneficial for a downstream task.

\vspace{1.2mm}
\noindent \textbf{Neural vs.~Rule-Based Systems.} Neural systems underperform their rule-based counterparts on most facets. This gap is most pronounced on BenchIE-E, whereas it is much smaller on BenchIE-C: these observations strongly indicate that neural systems struggle the most with correct distribution of tokens across the (adjacent) extraction slots. 
They also do not attempt to remove the optional (i.e., unnecessary) tokens, as indicated by extremely low performance on BenchIE-M. On CaRB, however, these same neural systems yield the best performance. Being trained and validated on datasets with extractions similar to CaRB's, neural extractors seem to overfit to CaRB evaluation. Our fact-based multi-faceted evaluation, however, reveals that their extractions are far less likely to be useful down the stream. 


\section{Conclusion}

We introduced BenchIE: a benchmark for more reliable fact-level evaluation of OIE systems for English, Chinese and German. Unlike existing benchmarks, BenchIE takes into account fact-level equivalence of extractions:~it consists of \textit{fact synsets} that contain \textit{all} acceptable surface forms of the same fact. Further, \en BenchIE is multi-faceted -- it allows to evaluate OIE extractions w.r.t. several aspects relevant in common downstream tasks. Our experiments show that current benchmarks, with incomplete gold standard and approximate token-level matching, drastically overestimate fact extraction abilities of OIE systems. 
Currently, the limits of BenchIE are its relatively small size (300 sentences v.s.~CaRB's 1,200) and its time-consuming annotation process. A promising research direction is the investigation of trade-off between the manual effort and completeness of different OIE annotation strategies. In this scenario, BenchIE is an ideal point of reference: it can precisely quantify the completeness of some larger (non-exhaustive) OIE dataset created with limited or no manual effort.

\clearpage

\bibliography{main}
\bibliographystyle{acl_natbib}

\clearpage
\appendix
\section{Appendix: Annotation Guidelines}

\subsection{Annotation Guidelines for English}
\label{app:annotation_guidelines_en}

\subsubsection{General Principle} 
The annotator should manually extract verb-mediated triples from a natural language sentence. Each triple should represent two entities or concepts, and the verb-mediated relation between them. For example, from the input sentence \emph{"Michael Jordan, who is a former basketball player, was born in Brooklyn."}, there are three entities and concepts---\emph{Michael Jordan, former basketball player} and \emph{Brooklyn}---which are related as follows: \emph{("Michael Jordan"; "is"; "former basketball player")} and \emph{("Michael Jordan"; "was born in"; "Brooklyn")}.

Once the triple is manually extracted, it should be placed into the correct fact synset (see Section~\ref{app:fact-synset}).

\subsubsection{Fact Synsets}\label{app:fact-synset}
Once a triple is manually extracted, the annotator should place the triple into its corresponding fact synset (for more details about the concept of fact synsets, refer to Section~\ref{sec:benchie}). In case there is no existing fact synset for the manually extracted triple, the annotator should create one and place the triple in that synset. 

\paragraph{Coreference.} The annotator should place extractions that refer to the same entity or concept under the same fact synset. Consider the following input sentence: \emph{"His son , John Crozie, was an aviation pioneer."}; the following triples should be placed in the same fact synset:
\begin{itemize}
    \item \emph{("His son"; "was"; "[an]\footnote{words in square brackets indicate optional tokens (see Section \ref{app:subsec-opt-tokens})} aviation pioneer")}
    \item \emph{("J.~Crozie"; "was"; "[an] aviation pioneer")}
\end{itemize}
because \emph{"His son"} and \emph{"John Crozie"} refer to the same entity.

\paragraph{Token placements within the slots.} The annotator should consider placing certain tokens in different slots, without damaging the meaning of the fact. 
Consider the input sentence \emph{"Michael Jordan was born in Brooklyn."}. There is one fact synset ($f_1$) and its corresponding triples ($t_1, t_2$ and $t_3$):
\begin{enumerate}
    \item[$f_1$] \emph{$t_1:$ ("M.~J."; "was born in"; "Brooklyn")} \\
                 \emph{$t_2:$ ("M.~J."; "was born"; "in Brooklyn")} \\
                 \emph{$t_3:$ ("M.~J."; "was"; "born in Brooklyn")}
\end{enumerate}
In $t_1$, the preposition \emph{"in"} is in the relation, while in $t_2$ it is in the object. Likewise, the annotator should allow for some flexibility w.r.t.~the verbs. While the verbs and prepositions naturally belong to the relation, some OIE systems were designed with different goal in mind; e.g., to detect head verbs as relations for detecting clauses within the extractions \cite{del2013clausie} or to fit SRL  frames for predicates \cite{stanovsky2018supervised}. We do not want to penalize the OIE systems for such design choices.

For BenchIE-E\footnote{For details on BenchIE-E, see Section \ref{subsec:benchie-ent}.}, however, this flexibility of token placements is not allowed. In particular, for $f_1$ the annotator is allowed to only extract $t_1$, while $t_2$ and $t_3$ should not be listed. Note that this is the only difference in the annotation guidelines between BenchIE-E and the standard BenchIE facet.

\paragraph{Passive voice.} When possible, if an extraction is in passive voice, the annotator should place its active voice equivalent into the appropriate fact synset. For instance, consider the sentence \emph{"The ball was kicked by John."}; then, the fact synset should contain the following triples:
\begin{itemize}
    \item \emph{("[The] ball"; "was kicked by"; "John")}
    \item \emph{("John"; "kicked"; "[The] ball")}
\end{itemize}
Note that the opposite direction is not allowed. If the sentence was \emph{"John kicked the ball."}, then the annotator is not allowed to manually extract the triple \emph{("[The] ball"; "was kicked by"; "John")} because such extraction contains words that are not originally found in the input sentence (\emph{"was"} and \emph{"by"}). These are so-called implicit extractions and we do not consider them (for details, see Section~\ref{app:subsec-implicit} of the appendix).

\subsubsection{Optional Tokens} 
\label{app:subsec-opt-tokens}
If possible, the annotator should label as \emph{optional} all tokens that can be omitted in an extraction without damaging its semantics.
Such tokens include determiners (e.g., \emph{a, the, an}), honorifics (e.g., \emph{[Prof.]~Michael Jordan}) or certain quantities (e.g., 
\emph{[some] major projects}. The optional tokens are marked with square brackets $[~]$. In what follows, we show examples of considered optional token(s).

\paragraph{Determiners.} Unless a determiner is a part of a named entity (e.g., \emph{"The Times"}), it is considered as optional. For instance, the following
triples are considered to be semantically equivalent: 
\begin{itemize}
    \item \emph{("Michael Jordan"; "took"; "the ball")}
    \item \emph{("Michael Jordan"; "took"; "ball")}
\end{itemize}
The annotator, therefore, should annotate \emph{("Michael Jordan"; "took"; "[the] ball")}, where the optional token is in square brackets.

\paragraph{Titles.} Titles of people are considered optional; e.g., \emph{("[Prof.] Michael Jordan"; "lives in"; "USA")}.

\paragraph{Adjectives.} The annotator should label adjectives as optional if possible. For example, in the following triple, the adjective \emph{"smart"} can be considered optional: \emph{("Albert Einstein"; "was"; "[a] [smart] scientist")}. Note that the annotator should be careful not to label adjectives as optional if they are essential to the meaning of the triple. For instance, the adjective \emph{"cold"} should not be labeled as optional in the triple \emph{("Berlin Wall"; "is [infamous] symbol of"; "[the] cold war")}. 

\paragraph{Quantities.} Certain quantities that modify a noun phrase can be considered as optional; e.g., \emph{("Mitsubishi"; "has control of"; "[some] major projects")}.

\paragraph{Words indicating some tenses.} The annotator can treat certain verbs that indicate tense as optional. For instance, the word \emph{"has"} in \emph{("FDA"; "[has] approved"; "Proleukin")} can be considered as optional, since both VPs \emph{"have approved"} and \emph{"approved"} contain the same core meaning. 

\paragraph{Verb phrases.} It is allowed for the annotator to mark verb phrases as optional if possible; e.g. \emph{("John"; "[continues to] reside in";
"Berlin")}.


\subsubsection{Attribution Clauses} Extractions that indicate attribution of another core piece of information should be placed in separate fact synset, because they indicate a separate piece of information with separate predicate. For example, the core information of the sentence \emph{"Conspiracy theorists say that Barack Obama was born in Kenya."} is that Barack Obama was born in Kenya. As indicated by \citet{mausam2012open}, it is important for OIE systems to extract the context about the attribution of such information. Therefore, the annotator should extract the core information---the triple \emph{("Barack Obama"; "[was] born in"; "Kenya")}---in one fact synset, and the triples indicating attribution---\emph{("Conspiracy theorists"; "say that"; "Barack Obama was born in Kenya")}---in another.

\subsubsection{Incomplete Clauses} 
The annotator should not extract incomplete clauses, i.e., triples that lack crucial piece of information. Suppose there is the input sentence \emph{"He was honored by the river being named after him".} The following triple should not be manually extracted: \emph{("He"; "was honored by"; "[the] river")}, but the following triples should be: \emph{("He"; "was honored by [the] river being named after"; "him")} and \emph{("[the] river"; "being named after"; "him")}.

\subsubsection{Overly Complex Extractions}
The annotators should not manually extract overly specific triples, such that their arguments are complex clauses. For instance, for the input sentence 
\emph{"Vaccinations against other viral diseases followed, including the successful rabies vaccination by Louis Pasteur in 1886."}, the following triple
should not be extracted: \emph{("Vaccinations against other viral diseases"; "followed"; "including the successful rabies vaccination by Louis Pasteur in 1886")} because the object is a complex clause which does not describe a single concept precisely, but rather it is composed of several concepts.

\subsubsection{Conjunctions}
The annotator should not allow for conjunctive phrases to form an argument (i.e., subject or object). Such arguments should be placed into separate extractions (and in separate fact synsets). Consider the sentence \emph{"Michael Jordan and Scottie Pippen played for Chicago Bulls.".} The annotator should manually extract the following triples:
\begin{itemize}
    \item \emph{("M.~Jordan"; "played for"; "Chicago Bulls")}
    \item \emph{("S.~Pippen"; "played for"; "Chicago Bulls")}
\end{itemize}
The annotator should not, however, extract \emph{("M.~J.~and S.~P."; "played for"; "Chicago Bulls")}.

\subsubsection{Implicit Extractions} 
\label{app:subsec-implicit}
We focus on explicit extractions, which means that every word in the extracted triple must be present in the original input sentence. Therefore, implicit extractions---i.e., extractions that contain inferred information with words not found in the sentence---are not considered. One example implicit extraction is \emph{("Michael Jordan"; "be"; "Prof.")} from the input sentence \emph{"Prof.~Michael Jordan lives in USA."}, where the triple infers that Michael Jordan is professor without being explicitly indicated in the sentence (i.e., the word \emph{"be"} is not present in the input sentence, it is inferred).




\subsection{Annotation Guidelines (Chinese)}
\label{app:annotation_guidelines_zh}

The annotator should follow the same general principles as with the English annotation guidelines (Section~\ref{app:annotation_guidelines_en}). 
Due to the language difference, we slightly adapted the annotation guidelines for the Chinese language. In what follows, we list those differences.

\subsubsection{Articles}
Chinese language does not contain articles (i.e., \emph{"a", "an", "the")}. Therefore, in the manual translation of the sentences, there are no articles in the Chinese counterparts. 

\subsubsection{Prepositional Phrases within a Noun Phrase}
Certain noun phrases with nested prepositional phrase cannot be translated directly into Chinese the same way as in English. For example, suppose we have
the phrase \emph{"Prime Minister of Australia"}. In Chinese, the literal translation of this phrase would be \emph{"Australia's Prime Minister"}. For instance,  
in the English annotations the sentence \emph{"He was the Prime Minister of Australia"} would have two fact synsets:
\begin{itemize}
    \item[$f_1$] \emph{("He"; "was [the] Pr.~Min.~of"; "Australia")}
    \item[$f_2$] \emph{("He"; "was"; "[the] Pr.~Min.~[of Australia]")}
\end{itemize}
This is because the fact synset $f_1$ relates the concepts \emph{"he"} and \emph{"Australia"} with the relation \emph{"was [the] Prime Minister of"}, while the second 
fact synset relates the concepts \emph{"he"} and \emph{"Prime Minister [of Australia]"} with the relation \emph{"was"}. 

In Chinese language, however, the construction of $f_1$ would not be possible, because the phrase \emph{"Prime Minister of Australia"} 
cannot be separated into \emph{"Prime Minister"} and \emph{"Australia"}. Therefore, the golden annotation for this particular example in Chinese would 
be only one fact synset: \emph{("He"; "was"; "[Australia's] Prime Minister")}, which is equivalent with $f_2$. 


\subsection{Annotation Guidelines (German)}
\label{app:annotation_guidelines_de}

In general, the annotators for German should follow the same guidelines described in Section \ref{app:annotation_guidelines_en} for English. In what follows, we describe the differences which are specific for the German annotations.


\subsubsection{Separable Verbs}
Separable verbs (e.g., \emph{"aufstehen"}) in German consist of a lexical core (a verb; \emph{"stehen"}) and a separable particle (e.g., a preposition; \emph{"auf"}). When used in a sentence, separable verbs in German are split in such manner that the separable particle goes to the end of the sentence. Consider the following sentence that contains the separable verb \emph{"\underline{aufstehen}"}: \emph{"Ich \underline{stehe} um 7 Uhr \underline{auf}"}. To accommodate the verb-mediated relations, the annotator should extract the separable particle right after the separable core within the predicate: \emph{("Ich"; "\underline{stehe auf} um"; "7 Uhr")}

\subsubsection{Modal Verbs} 
The modal verbs follow similar pattern as the separable verbs. Namely, the modal verb has the main predicate position within the sentence (directly followed by the subject), and the main verb that is modified by the modal verb is at the end of the sentence; e.g.~sentence \emph{"I \underline{must go} to work"} and its German counterpart \emph{"Ich \underline{muss} zur Arbeit \underline{gehen}"}. Following the same guidelines for verb-mediated predicates, the annotator should extract the modal verb together with the main verb: \emph{("Ich"; "\underline{muss gehen} zur"; "Arbeit")}.

\subsubsection{Passive Voice}
Consider the following English sentence written in passive voice \emph{"The letters \underline{were sent} through the messenger"} and its German counterpart \emph{"Die Briefe \underline{wurden} durch den Boten \underline{geschickt}"}. Following the spirit of extractions with verb-mediated relations, the annotator should extract the following triple: \emph{("[Die] Briefe"; "\underline{wurden geschickt} durch"; "[den] Boten")}.






\section{Annotation Tool}
\label{app:annotation-tool}

\begin{figure*}
    \centering
    \includegraphics[scale=0.7]{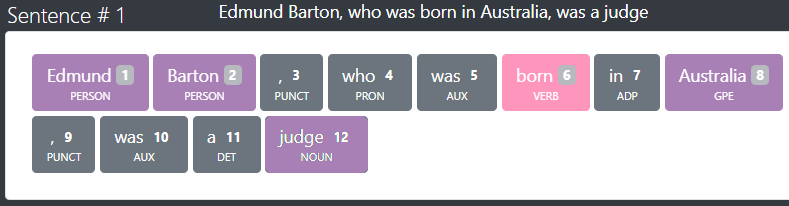}
    \caption{Highlighting tokens of interest: verbs (potential relations) and nouns (potential arguments).}
    \label{fig:annie-highlight}
\end{figure*}

\begin{figure*}
    \centering
    \includegraphics[scale=0.7]{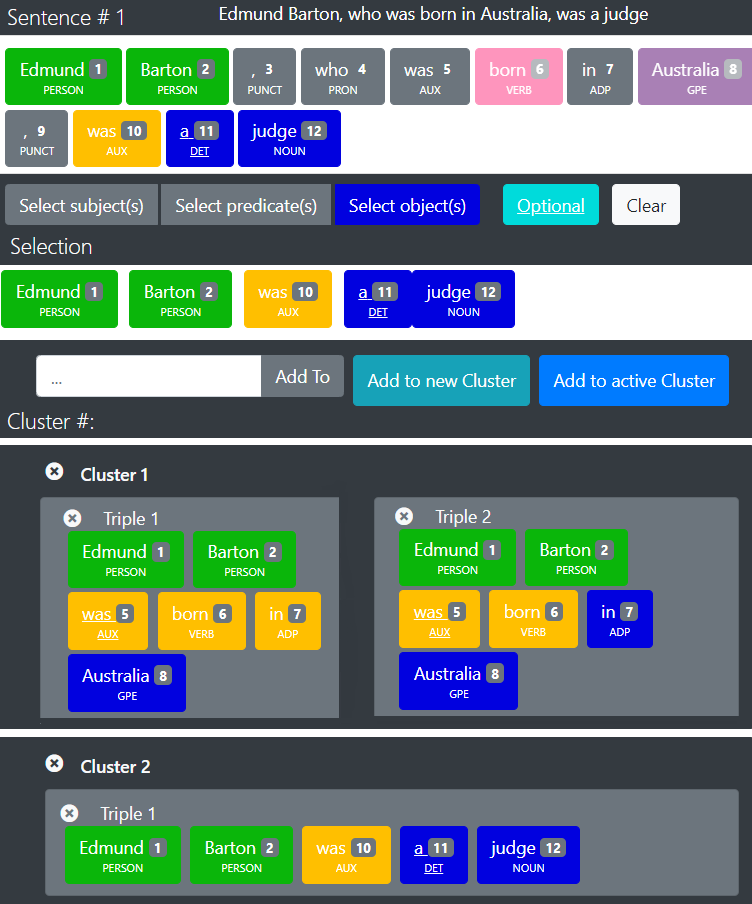}
    \caption{Manual labeling of OIE triples. The user selects tokens from the tokenized input sentence and 
    places them into the correct slot: \textcolor{forestgreen}{subject (green)}, \textcolor{orange}{predicate (yellow)} or \textcolor{blue}{object (blue)}. Then, the user adds the extracted triple either to an active fact cluster (i.e., fact synset) or to a new one. The user can also select which tokens are optional by clicking the "Optional" button on an active token selection.}
    \label{fig:annie-fact-label}
\end{figure*}

To facilitate the annotation process, we developed a web-based annotation tool: AnnIE \cite{friedrich2021annie}. First, the annotator is given the input sentence as a string along with its tokenized form (Figure~\ref{fig:annie-highlight}). Then, the tool highlights the tokens of interest that are candidates for the slots. In particular, we highlight the verbs in one color (candidate predicates) and the nouns in another (candidate arguments). 

\begin{figure*}
    \centering
    \includegraphics[scale=0.6]{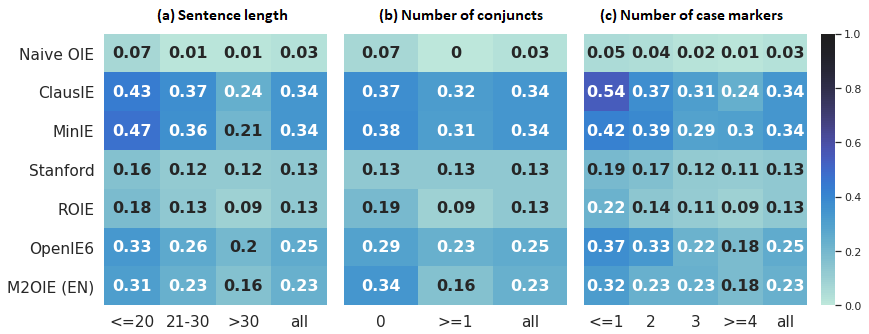}
    \caption{Bucketized experiments: F1 score according to different bucketizations of the input sentences: sentence length (a); number of conjunctions (b); number of case markers (c).}
    \label{fig:buckets}
\end{figure*}

Then, the annotator can select the tokens with a UI and place them into slots. This forms one annotated triple. Note that the annotator can also annotate for optional tokens and phrases with the use of the mouse double click. Then, the annotator can place the newly annotated triple in either a new fact synset (cluster) or in an existing one (Figure~\ref{fig:annie-fact-label}). For more details on the annotation tool, see \cite{friedrich2021annie}.

\section{Further Error Analysis}
\label{app:bucket-experiments}

\label{app:bucketization}

Based on preliminary qualitative error analysis, we chose bucketization according to some linguistic properties of the sentences that produced erroneous triples. In particular, we examine the performance of OIE systems for sentence length, presence of conjunctions and case markers, since these appeared to be the most common reasons for failure. Note that BenchIE allows for any type of bucketization, which can be used for diverse set of fine-grained evaluation for future research on OIE. 

\subsection{Sentence Length}

Sentence length is a feature that can affect the performance of NLP systems for different tasks, including relation extraction \cite{alt2020probing} and named entity recognition \cite{arora2021identifying}. To evaluate how sentence length afffects performance of OIE systems as well, we split the sentences into three buckets: sentences shorter or equal than 20 tokens, between 21 and 30 tokens, and more than 30 tokens. The distribution of these buckets are 120, 113 and 67 sentences respectively. 

We observed that shorter sentences usually yield the best performance for all OIE systems w.r.t.~the $F_1$ score (Figure~\ref{fig:buckets}a). An extreme example is MinIE, which loses 26 percentage points from sentences shorter than 20 tokens to sentences longer than 30 tokens. Part of the reason why such sentences are harder to handle is because they contain more complex linguistic structures, such as conjunctions and case markers. Such sentences tend to to produce overly complex extractions that contain very complex structures in their arguments (see example extraction $t_3$ in Table \ref{tab:common_errors}).

\setlength{\tabcolsep}{3pt}
\begin{table*}
    \centering
    \footnotesize
        \begin{tabular}{rrclr}  
            \toprule
            Extraction ID &   \multicolumn{3}{c}{Extractions}     & BenchIE \\
            \cmidrule(r){1-5}
            
            \multicolumn{5}{l}{\textbf{Sentence $s_1$:} \emph{"A large gravestone was erected in 1866 , over 100 years after his death."}} \\
            \midrule 

        $t_1$ & \emph{("A large gravestone";} & \emph{"was erected";} & \emph{"in 1866 over 100 y.~after his death")} &  0 \\ 
               $t_2$ & \emph{("A large gravestone";} & \emph{"was erected";} & \emph{"in 1866")} &  1 \\
            \cmidrule(r){1-5}
            \multicolumn{5}{l}{\textbf{Sentence $s_2$:} \emph{"The brightest star in Serpens, Alpha Serpentis, or Unukalhai, is a red giant of spectral type K2III}}  \\ 
             \multicolumn{5}{l}{\emph{\hspace{1.9cm}located approximately away which marks the snake's heart ."}} \\
                         \cmidrule(r){1-5}

               $t_3$ & \emph{("The brightest star in Serpens,} &	\emph{"is"}	& \emph{"a red giant of sp.~type K2III loc. app.}  &  0 \\
               & \emph{Alpha Serpentis , or Unukalhai"} & & \emph{away which marks the snake 's heart")} \\
               $t_4$ & \emph{("brightest star in Serpens";}	& \emph{"is";} & \emph{"red giant")} &  1 \\ \hline

            
              \multicolumn{5}{l}{\textbf{Sentence $s_3$:} \emph{"Lugo and Lozano were released in 1993 and continue to reside in Venezuela.}} \\ 
              \cmidrule(r){1-5}
             $t_5$   & \emph{("Lugo and Lozano";} &	\emph{"released";}	& \emph{"in 1993")} &  0 \\
              $t_6$ & \emph{("Lugo";}                 &  \emph{"were released";}     &   \emph{"in 1993")} &  1 \\
              $t_7$ &		    \emph{("Lozano";}                 &  \emph{"were released";}     &   \emph{"in 1993")} &  1 \\
            \bottomrule
        \end{tabular}
    \caption{Example extractions along with their score on BenchIE.} 
    \label{tab:common_errors}
\end{table*}

\subsection{Conjunctions}
To examine the effect of the conjunctions on the performance of OIE systems, we bucketized the input sentences according to the dependency type \texttt{conj}, which relates two conjunct words in a sentence. In particular, we place the sentences with no conjuncts in one bucket, and the sentences with one or more conjuncts in another bucket. With such bucketization, half of the sentences are in the first bucket, and half in the other. We observed that the F1 score suffers when a sentence contains at least one pair of conjuncts (Figure \ref{fig:buckets}b). This observation partially explains the observation from Section~\ref{subsec:slot-errors}  that OIE systems have troubles identifying the objects correctly. In subsequent experiments, we observed that sentences with more than one conjuncts worsen the scores further compared to the sentences with one or no conjuncts. The triple $t_5$ in Table \ref{tab:common_errors} is an example of such erroneous extraction.

Neural models seem to suffer the most due to the conjuncts. For instance, M$^2$OIE loses more than half of the F1 score points (from 0.34 down to 0.16) when at least one conjunct is found in the sentence. The exception for the neural systems is OpenIE 6, which is more stable (goes down from 0.29 to 0.23). The reason is because OpenIE 6 was specifically trained to handle conjunctions. Interestingly, ClausIE and MinIE---rule-based systems---lose approximately the same amount of F1 score points as the neural OpenIE 6. This indicates that neural models can be trained to handle conjunctions similarly as rule-based systems, though there is still room for improvement. We observed similar behaviors for coordinated conjunctions.

\subsection{Case Markers}
%
In preliminary qualitative experiments, we found that the objects are often overly specific because they include phrases that should in principle not be part of the expressed concept. Such excessively specific phrases are usually prepositional phrases or case markers. Consider, for example, the triple $t_1$ in Table~\ref{tab:common_errors}. The object in this triple is overly specific and, thus, incorrect. 

To quantify the effect of such case markers, we bucketized the data according to the number of the typed dependencies \texttt{case} that are found in the input sentences. We observed that, as the number of \texttt{case} dependencies increases, the performance of OIE systems  decreases (Figure~\ref{fig:buckets}c). We observed similar behavior for the number of prepositions in a sentence.
The rule-based system ClausIE is very sensitive w.r.t.~this property, while MinIE is more stable. MinIE was built on top of ClausIE and also focused on restructuring the output of ClausIE, which is the likely reason why MinIE is more robust w.r.t.~the case markers. Neural systems (ROIE, OpenIE 6 and M$^2$OIE) are very sensitive to this property, since their performance is much lower when we compare the buckets of 0 or 1 case dependency and the buckets with more than 4 case dependencies.

\section{More Detailed Discussion on Related Work}
\label{sec:rw_ext}


\subsection{OIE Benchmarks}
The currently existing benchmarks are based on token-based scoring. The first attempt to create an OIE benchmark was OIE2016 \cite{Stanovsky2016EMNLP}. The authors used a dataset from another task---QA-SRL \cite{he2015question}---and automatically ported it to OIE. For scoring an OIE triple, they follow the original task's guidelines \cite{he2015question} and match only the grammatical heads of each slot from the OIE triple with the ones from the golden datasets. Such approach has many drawbacks \cite{zhan2020span}, because (1) every error in the automatic porting transfers over to the evaluation dataset; (2) triples are incorrectly (and over-optimistically) scored because it only considers token-overlaps on grammatical heads, not the whole slots. Being crowdsourced, CaRB \cite{bhardwaj2019carb} improves over OIE2016 by aggregating per-slot token-level precision and recall scores between system and gold extractions across the three slots (subject, predicate, and object). However, such approach is overly-lenient, as it allows for incorrect extractions to be scored positively (see examples in Table \ref{tab:carb-benchie-extractions2}). Subsequent work followed similar evaluation procedures. For instance, \citet{dong2021docoie} propose a dataset that evaluates  document-level OIE which uses the same scoring procedures as CaRB. 

\subsection{Multi-faceted Evaluation}
While having a reliable single-metric benchmark is crucial for the progress of NLP, recent research indicated that focusing on single metrics is somewhat limited, because it does not provide further insights that go beyond the averaged scores \cite{ethayarajh2020utility,narayan2021personalized}. In particular, \citet{ethayarajh2020utility} argue that single-metric scores ignore certain properties of the evaluated NLP models. Such properties, however, could be relevant for practitioners or for certain downstream tasks. As a consequence, the final evaluation score is computed at the expense of other properties of the model. To allow such multi-faceted evaluations, \citet{liu2021explainaboard} proposed ExplainaBoard, which scores NLP systems from several tasks across different facets, and \citet{vath2021beyond} propose a multi-faceted benchmark for visual question answering. 

Due to the incompleteness of current OIE benchmarks---and because of the peculiarity of the task---no such multi-faceted evaluation for OIE has been proposed. For each tested extraction, the state-of-the-art benchmarks provide scores that are in the interval of $[0, 1]$. Such design is employed because the benchmarks are incomplete, which, in turn, makes it difficult to do proper multi-faceted evaluation. To tackle this issue, we propose a multi-faceted evaluation that scores OIE systems across several facets that are important for downstream tasks (see details in Section \ref{sec:facets}).

\subsection{Automatic Error Analysis}
Producing automatic error analysis with current benchmarks is not trivial because they are not exhaustive and do not provide crisp scores. For instance, when there are scores within the interval of $[0,1]$ for each slot---as in CaRB---, it is hard to say where exactly the error occurred. Previous work on OIE performed error analysis manually \cite{fader2011identifying,schneider2017relvis}, which is very time-consuming and inefficient. In contrast to prior work, BenchIE is exhaustive benchmark that provides crisp scores, which allows for automatic per-slot error analysis. We discuss BenchIE's automatic error analysis approach in Section \ref{sec:profiling}.


\end{document}